\documentclass[12pt,a4paper]{article}

\usepackage{graphicx}
\usepackage{amsmath}
\usepackage{amssymb}
\usepackage[font=small]{caption}
\usepackage{fullpage}
\usepackage[natbibapa]{apacite} 
\usepackage{color}

\usepackage{setspace}
\usepackage[T1]{fontenc}
\usepackage[utf8]{inputenc}
\usepackage{authblk}
\usepackage{url}

\usepackage{soul,color}
\usepackage{float}

\title{Corpus specificity in LSA and Word2vec: the role of out-of-domain documents}

\author[1,2]{Edgar Altszyler \thanks{Corresponding author: ealtszyler@dc.uba.ar}}
\author[3]{Mariano Sigman}
\author[1,2]{Diego Fern\'andez Slezak}

\affil[1]{ Universidad de Buenos Aires. Facultad de Ciencias Exactas y Naturales. Departamento de Computaci\'on. Buenos Aires, Argentina.}
\affil[2]{Instituto de Investigaci\'on en Ciencias de la Computaci\'on, UBA-CONICET}
\affil[3]{Universidad Torcuato Di Tella - CONICET.}

\date{\vspace{-5ex}}
 
\begin{document}

\maketitle

%\begin{Abstract}
Latent Semantic Analysis (LSA) and Word2vec are some of the most widely used word embeddings. Despite the popularity of these techniques, the precise mechanisms by which they acquire new semantic relations between words remain unclear. In the present article we investigate whether LSA and Word2vec capacity to identify relevant semantic dimensions increases with size of corpus. One intuitive hypothesis is that the capacity to identify relevant dimensions should increase as the amount of data increases. However, if corpus size grow in topics which are not specific to the domain of interest, signal to noise ratio may weaken. Here we set to examine and distinguish these alternative hypothesis. To investigate the effect of corpus specificity and size in word-embeddings we study two ways for progressive elimination of documents: the elimination of random documents vs. the elimination of documents unrelated to a specific task. We show that Word2vec can take advantage of all the documents, obtaining its best performance when it is trained with the whole corpus. On the contrary, the specialization (removal of out-of-domain documents) of the training corpus, accompanied by a decrease of dimensionality, can increase LSA word-representation quality while speeding up the processing time. Furthermore, we show that the specialization without the decrease in LSA dimensionality can produce a strong performance reduction in specific tasks. From a cognitive-modeling point of view, we point out that LSA's word-knowledge acquisitions may not be efficiently exploiting higher-order co-occurrences and global relations, whereas Word2vec does.
%\end{Abstract}

\section{Introduction}

The main idea behind corpus-based semantic representation is that words with similar meanings tend to occur in similar contexts \citep{Harris1954}. 
This proposition is called \textit{distributional hypothesis} and provides a practical framework to understand and compute semantic relationship between words. 
Based in the \textit{distributional hypothesis}, Latent Semantic Analysis (LSA) \citep{deerwester1990indexing,Landauer1997,Hu2007} and Word2vec \citep{Mikolov2013a,Mikolov2013b,church_2017}, are one of the most important methods for word meaning representation, which describes each word in a vectorial space, where words with similar meanings are located close to each other.

Word embeddings have been applied in a wide variety of areas such as information retrieval \citep{deerwester1990indexing}, psychiatry \citep{Bedi2015}, literature \citep{altszyler2015analisis}, education\citep{Jorge-Botana2010} and cognitive sciences \citep{Landauer1997, Denhiere2004,Lemaire2004,Diuk2012}.

LSA takes as input a training Corpus formed by a collection of documents. 
Then a word by document co-occurrence matrix is constructed, which contains the distribution of occurrence of the different words along the documents. 
Then, usually, a mathematical transformation is applied to reduce the weight of uninformative high frequency words in the words-documents matrix  
\citep{Dumais1991}. 
Finally, a linear dimensionality reduction is implemented by a truncated \textit{Singular Value Decomposition}, SVD, which projects every word in a subspace of a predefined number of dimensions, \textit{k}. 
The success of LSA in capturing the latent meaning of words comes from this low-dimensional mapping. 
This representation improvement can be explained as a consequence of the elimination of the most noisy dimensions \citep{Turney2010}. 

Word2vec consists of two neural network models, Continuous Bag of Words (CBOW)
and Skip-gram. To train the models, a sliding window is moved along the corpus. In the CBOW scheme, in each step the neural network is trained to predict the center word (the word in the center of the window based) given the context words (the other words in the window). While in the skip-gram scheme, the model is trained to predict the context words based on the central word. In the present paper we use the skip-gram, which has produced better performance in \citep{Mikolov2013b}.

Despite the development of new word representation methods, LSA is still intensively used, and has been shown that produce better performances than Word2vec methods in small to medium size training corpus \citep{altszyler2017interpretation}.

\subsection*{Training Corpus Size and Specificity in Word-embeddings}\label{intro_corpus}
Over the last years, great effort has been devoted to understand how to choose the right parameter settings for different tasks \citep{QuesadaCreating,Dumais2003,Landauer1997,lapesa2014large,Bradford2008,Nakov2003,baroni2014don}. 
However, considerably lesser attention has been given to study how different corpus used as input for training may affect the performance. 
Here we ask a simple question on the property of the corpus: is there a monotonic relation between corpus size and the performance? 
More precisely, what happens if the topic of additional documents differ from the topics in the specific task? 
Previous studies have surprisingly shown some contradictory results on this simple question. 

On the one hand, in the foundational work, Landauer \emph{et} al. \citep{Landauer1997} compare the word-knowledge acquisition between LSA and that of children's.
This acquisition process may be produced by 1) direct learning, enhancing the incorporation of new words by reading texts that explicitly contain them; or 2) indirect learning, enhancing the incorporation of new words by reading texts that do not contain them.
To do that, they evaluate LSA semantic representation trained with different size corpus in multiple-choice synonym questions extracted from the TOEFL exam.
This test consists in 80 multiple-choice questions, in which its requested to identify the synonym of a word between 4 options. 
In order to train the LSA, Landauer and Dumais used the TASA corpus \citep{TASA}. 

Landauer \emph{et} al. \citep{Landauer1997}  randomly replaced exam-words in the corpus with non-sense words and varied the number of corpus' documents selecting nested sub-samples of the total corpus. 
They concluded that LSA improves its performance on the exam both when training with documents with exam-words and without them.
However, as could be expected, they observed a grater effect when training with exam-words. 
It is worth mentioning that the replacement of exam-words with non-sense words may create incorrect documents, thus, making the algorithm acquire word-knowledge from documents which should have an exam-word but do not.
In the Results section, we will study this indirect word acquisition in the TOEFL test without using non-sense words.

Along the same line, \citep{Lemaire2006} studied the effect of high-order co-occurrences in LSA semantic similarity, which goes further in the study of Landauer's indirect word acquisition.

In their work, Landauer \emph{et} al. \citep{Lemaire2006} measure how the similarity between 28 pairs of words (such as bee/honey and buy/shop) changes when a 400-dimensions LSA is trained with a growing number of paragraphs. 
Furthermore, they identify for this task the marginal contribution of the first, second and third order of co-occurrence as the number of paragraphs is increased. 
In this experiment, they found that not only does the first order of co-occurrence contribute to the semantic closeness of the word pairs, but also the second and the third order promote an increment on pairs similarity. 
It is worth noting that Landauer's indirect word acquisition can be understood in terms of paragraphs without either of the words in a pair, and containing a third or more order co-occurrence link.

So, the conclusion from the Lemaire \emph{et} al. \citep{Lemaire2006} and Landauer \emph{et} al. \citep{Landauer1997} studies suggest that increasing corpus size results in a gain, 
even if this increase is in topics which are unrelated for the relevant semantic directions which are pertinent for the task. 

However, a different conclusion seems to result from other set of studies. \cite{Stone2006} have studied the effect of Corpus size and specificity in a document similarity rating task. 
They found that training LSA with smaller subcorpus selected for the specific task domain maintains or even improves LSA performance. This corresponds to the intuition
of noise filtering, when removing information from irrelevant dimensions results in improvements of performance. 

In addition, Olde \emph{et} al. \citep{Olde2002} have studied the effect of selecting specific subcorpus in an automatic exam evaluation task. 
They created several subcorpus from a Physics corpus, progressively discarding documents unrelated to the specific questions. 
Their results showed small differences in the performance between the LSA trained with original corpus and the LSA trained with the more specific subcorpus. 

It is well known that the number of LSA dimensions (\textit{k}) is a key parameter to be duly adjusted in order to eliminate the most noisy dimensions \citep{Landauer1997,Turney2010}. 
Excessively high \textit{k} values may not eliminate enough noisy dimensions, while excessively low k values may not have enough dimensions to generate a proper representation. 
In this context, we hypothesize that when out-of-domain documents are discarded, the number of dimensions needed to represent the data should be lower, thus, \textit{k} must be decreased. 

Regarding Word2vec, \cite{cardellinodisjoint} and \cite{dusserre2017bigger} have shown that Word2vec trained with a specific corpus can produce better performance in semantic tasks than when it is trained with a bigger and general corpus. Despite these works point out the relevance of domain-specific corpora, they do not study the specificity in isolation, as they compare corpus from different sources.

In this article, we set to investigate the effect of the specificity and size of training corpus in word-embeddings, and how this interacts with the number of dimensions.
To measure the semantic representations quality we have used two different tasks: the TOEFL exam, and a categorization test.
The corpus evaluation method consists in the comparison between two ways of progressive elimination of documents: the elimination of random documents vs the elimination of out-of-domain documents (unrelated to the specific task). 
In addition, we have varied \textit{k} within a wide range of values.

As we show, LSA's dimensionality plays a key role in the LSA representation when the corpus analysis is made. 
In particular, we observe that both, discarding out-of-domain documents and decreasing the number of dimensions produces an increase in the algorithm performance. 
In one of the two tasks, discarding out-of-domain documents without the decrease of \textit{k} results in the complete opposite behavior, showing a strong performance reduction. On the other hand, Word2vec shows in all cases a performance reduction when discarding out-of-domain, which suggests an exploitation of higher-order word co-occurrences.

Our contribution in understanding the effect of out-of-domain documents in  word-embeddings knowledge acquisitions is valuable from two different perspective:
\begin{itemize}
 \item From an operational point of view: we show that LSA's performance can be enhanced when: (1) its training corpus is \textit{cleaned} from  out-of-domain documents, and (2) a reduction of LSA's dimensions number is applied.
 Furthermore, the reduction of both the corpus size and the number of dimensions tend to speed up the processing time. On the other hand, word2vec can take advantage of all the documents, obtaining its best performance when it is trained with the whole corpus.
 
 \item From a cognitive modeling point of view: we point out that LSA's word-knowledge acquisition does not take advantage of indirect learning, while word2vec does. 
 This throws light upon models capabilities and limitations in modeling human cognitive tasks, such as: human word-learning \citep{Landauer1997,Lemaire2006,landauer2007}, semantic memory \citep{Denhiere2004,Kintsch2011,landauer2007} and words classification \citep{Laham1997}. 
\end{itemize}

\section*{Methods}

We used TASA corpus \citep{TASA} in all experiments. 
TASA is a commonly used linguistic corpus consisting of more than 37 thousand educational texts from USA K12 curriculum. 
We word-tokenized each document, discarding punctuation marks, numbers and symbols. 
Then, we transformed each word to lowercase and eliminated stopwords, using the stoplist in NLTK Python package \citep{Bird2009}. TASA corpus contains more than 5 million words in its cleaned version. 

In each experiment, the training corpus size was changed by discarding documents in two different ways:
\begin{itemize}
 \item \textit{Random documents discarding:} The desired number of documents (\textit{n}) contained in the subcorpus is preselected. Then, documents are randomly eliminated from the original corpus until there are exactly \textit{n} documents. 
 If any of the test words (i.e. words that appear in the specific task) does not appear at least once in the remaining corpus, one document is randomly replaced with one of the discarded documents that contains the missing word.   

 \item \textit{Out-of-domain documents discarding:} The desired number of documents (\textit{n}) contained in the subcorpus is preselected. 
 Then, only documents with no test words are eliminated from the original corpus until there are exactly \textit{n} documents. 
 Here, \textit{n} must be greater than or equal to the number of documents that contain at least one of the test words.
\end{itemize}
Both, LSA and Skip-gram word-embeddings were generated with Gensim Python library \citep{rehurek_lrec}. In LSA implementation, a Log-Entropy transformation was applied before the truncated Singular Value Decomposition. In Skip-gram implementation, we discarded tokens with frequency higher than $10^{-3}$, and we set the window size and negative sampling parameters to 15 (which were found to be maximal in two semantic tasks over TASA corpus \citep{altszyler2017interpretation}).
In all cases, word-embeddings dimensions values were varied to study its dependency. 

The semantic similarity ($S$) of two words was calculated using the cosine similarity measure between their respective vectorial representation ($\mathbf{v_1}$,$\mathbf{v_2}$),
\begin{equation}
S(\mathbf{v_1},\mathbf{v_2})=cos(\mathbf{v_1},\mathbf{v_2})=\frac{\mathbf{v_1}.\mathbf{v_2}}{\|\mathbf{v_1}\|.\|\mathbf{v_2}\|}
\end{equation}

The semantic distances between two words $d(\mathbf{v_1},\mathbf{v_2})$ is calculated as 1 minus the semantic similarity ( $d(\mathbf{v_1},\mathbf{v_2}) = 1- S(\mathbf{v_1},\mathbf{v_2}) $).

Word-embeddings knowledge acquisition was tested in two different tasks: a semantic categorization test and the TOEFL test.

\subsection*{Semantic categorization test}\label{method_categs}

In this test we measured the capabilities of the model to represent the semantic categories used by Patel \textit{et} al. \citep{Patel1997} (such as, drinks, countries, tools and clothes). 
The test is composed by 53 categories with 10 words each. 
In order to measure how well the word $i$ is grouped vis-\`{a}-vis the other words in its semantic category we used the Silhouette Coefficients, $s(i)$ \citep{Rousseeuw1987}, 

\begin{equation}
s(i) = \frac{b(i) - a(i)}{\max\{a(i),b(i)\}},
\end{equation}
where $a(i)$ is the mean distance of word $i$ with all other words within the same category, and $b(i)$ is the minimum mean distance of word $i$ to any words within another category (i.e. the mean distance to the neighbouring category). 
In other words, Silhouette Coefficients measure how close is a word to its own category words compared to the closeness to neighbouring words.
The Silhouette Score is computed as the mean value of all Silhouette Coefficients. 
The score takes values between -1 and 1, higher values reporting localized categories with larger distances between categories, representing better clustering.

The high number of test words (530) and the high frequency of some of them leaves only a few document with no test words.
This makes varied corpus size range in the \textit{out-of-domain documents discarding} very small. 
To avoid this, we tested only on the 10 least frequent categories. 
The frequency of a question is measured as the number of documents in which at least one word from this category appears.

\subsection*{TOEFL test}
The TOEFL test was introduced by \cite{Landauer1997} to evaluate the quality of semantic representations. 
This test consists of 80 multiple-choice questions, in which it is requested to identify the synonym of a target word between 4 options. 
For example: \textit{ select the most semantically similar to ``enormously'' between this words: ``tremendously'', ``appropriately'', ``uniquely'' and ``decidedly''}.
The performance of this test was measured by the percentage of correct responses.

Again, The high number of test words (400) and the high frequency of some of them leaves few documents with no test words.
So we performed the test only on the 20 least frequent questions in order to have out-of-domain documents to discard. 
\section*{Results}
\subsection*{Semantic categorization Test}
%We begin by measuring the categorization performance (Silhouette Score) in the 10 categories test set when the number of LSA's dimensions is varied (left panel of Figure\ref{siluette_vs_size}).
%It can be seen a bell shape with a maximum for 100 LSA's dimensions. Additionally, we plot in the right panel of Figure\ref{siluette_vs_size} the silhouette score dependence on the size of the corpus for the case of 100 dimensions. The corpus size variation was made by the elimination of random documents (Random documents discarding method).  As expected, the performance increases with the corpus size, given that there is more documents to extract useful relations.

%\begin{figure}[h!]
%\begin{center}
%\includegraphics[width=0.9\textwidth]{Fig1.eps}
%\caption{ {\bf Semantic categorization test performance.}
%Left panel: Performance (Silhouette Score) in function of the number of LSA's dimensions ($k$). Right panel: Silhouette Score in function of the corpus size for $k=100$. The corpus size variation was made discarding random documents. Each point represent 10 repetitions.}
%\label{siluette_vs_size}
%\end{center}
%\end{figure}

\begin{figure}[h!]
\begin{center}
\includegraphics[width=0.5\textwidth]{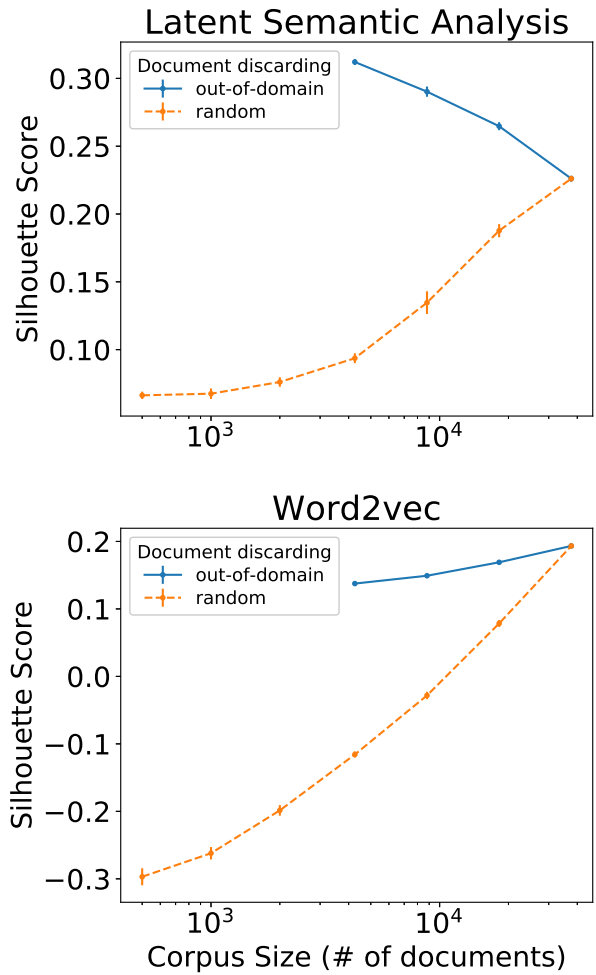}
\caption{Semantic categorization test analysis.
Silhouette Score vs corpus size for with both documents discarding methods: \textit{random document discarding} (dashed lines) and \textit{out-of-domain documents discarding} (solid lines). The shown Silhouette Score values and their error bars are, respectively, the mean values and the standard error of the mean of 10 samples. In most of the dots the error bars are not visible, this is because their length are smaller than the dot size. The dimension was varied among \{5, 10, 20, 50, 100, 300, 500, 1000\} for LSA and among  \{5, 10, 20, 50, 100, 300, 500\} for Word2vec. Due to the high computational effort, in the case of Word2vec we avoid using 1000 dimensions.}
\label{Silhuette_vs_size_low}
\end{center}
\end{figure}

In Figure \ref{Silhuette_vs_size_low} we show the LSA (top panel) and Word2vec (bottom panel) categorization performance with both documents discarding methods. For each corpus size and document discarding method we took 10 subcorpus samples (in total we consider 90 subcorpus + the complete corpus). In each corpus/subcorpus we trained LSA and Word2vec with a wide range of dimension values, using in each case the dimension that produces the best mean performance.

In both cases, performance decreases when documents are randomly discarded (dashed lines). However LSA and Word2vec have different behavior in the out-of-domain document discarding method (solid lines). While LSA produces better scores with increasing specificity, the word2vec performance decreases in the same situation.

%\begin{figure}[h!]
%\begin{center}
%\includegraphics[width=1\textwidth]{Fig2.eps}
%\caption{{\bf Semantic categorization test analysis.}
%Silhouette Score vs corpus size with both document variation methods: \textit{Random document discarding} (solid lines) and \textit{out-of-domain documents discarding} (dashed lines). The four graphs correspond to different LSA's dimensions values (k=20,50,100,300). The shown Silhouette Score values and their error bars are, respectively, the mean values and the standard error of the mean of 10 samples}
%\label{Silhuette_vs_size_low}
%\end{center}
%\end{figure}

LSA's maximum performance is obtained using 20 dimensions and removing all out-of-domain documents in the training corpus. While, when all the corpus is used the best number of dimensions is 100. These results show that performance for a specific task may be increased by ``cleaning'' the training corpus of out-of-domain documents. But, in order to enhance the performance, the elimination of out-of-domain documents should be accompanied by a decrease of the number of LSA dimensions. For example, fixing the number of dimensions to 100 the performance result in a reduction of 55\%. We also point out that this technical subtlety has not been taken into account in previous results that reported the presence of indirect learning in LSA \citep{Landauer1997,Lemaire2006}.

Additionally, the need of decreasing the LSA dimensionality when the corpus size is reduced only occurs in the \textit{out-of-domain documents discarding} method (see Figure \ref{S1_Fig} in Supplementary Materials). This result is consistent with LSA's ability to capture \emph{latent} semantic domains. Unlike \textit{random discarding} method, \textit{out-of-domain documents discarding} strongly reduces the topics variety, thus less dimensions are needed to identify the words categories. In contrast, Word2vec do not present a shift in its maximums in the \textit{out-of-domain documents discarding} method (see Figure \ref{S2_Fig} in Supplementary Materials). Moreover, Word2vec is little sensitive to changes in its dimensionality. These finding suggest that Word2vec do not encode \emph{latent} semantic domains, however more analysis must be done in these direction (see \citep{baroni2014don} discussion).

\subsection*{TOEFL Test}
\begin{figure}[h!]
\begin{center}
\includegraphics[width=0.5\textwidth]{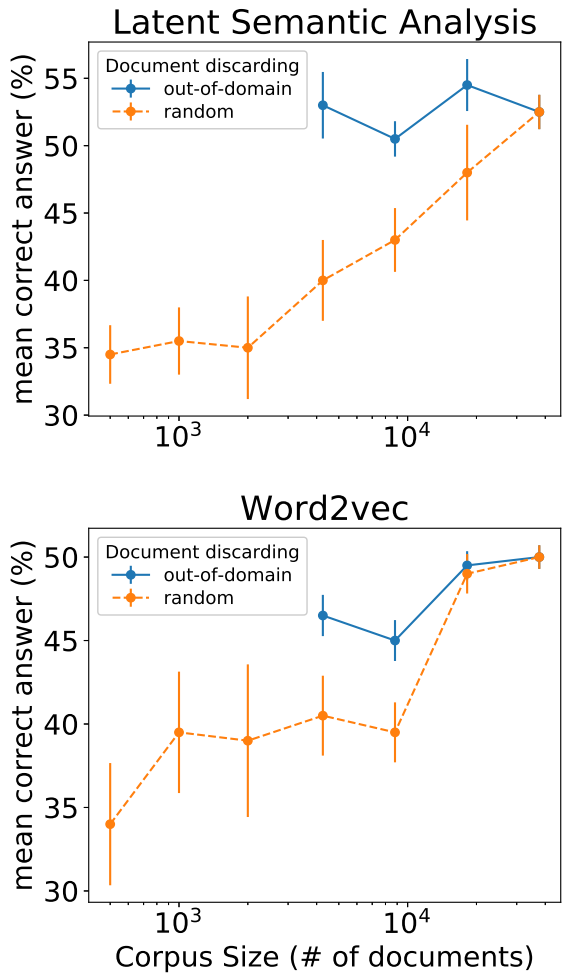}
\caption{TOEFL test analysis. Correct answer percentage vs corpus size with both document variation methods: \textit{Random document discarding} (dashed lines) and the \textit{out-of-domain documents discarding} (solid lines). The shown Silhouette Score values and their error bars are,respectively, the mean values and the standard error of the mean of 10 samples. The dimension was varied among \{5, 10, 20, 50, 100, 300, 500, 1000\} for LSA and among \{5, 10, 20, 50, 100, 300, 500\} for Word2vec. Due to the high computational effort, in the case of Word2vec we avoid using 1000 dimensions.}
\label{toefle_correct}
\end{center}
\end{figure}

In Figure\ref{toefle_correct} we show the TOEFL correct answer fraction vs the corpus size. We varied the corpus size by both methods:  the \textit{out-of-domain documents discarding} and the \textit{Random document discarding}. As in the categorization test procedure, a wide range of dimension values where tested, using in each case the dimension that produces the best mean performance. 

In both models, performance decreases when documents are randomly discarded (dashed lines in figure \ref{toefle_correct}). For LSA, the elimination of out-of-domain documents does not produce a significant performance variation, which shows that LSA can not take advantage of out-of-domain document. This results are in contradiction with \cite{Landauer1997} observation of indirect learning. We believe that this difference is due to the lack of adjustment in the number of dimensions. On the other hand, Word2vec has the same behaviour as in the categorization test. The performance when the out-of-domain documents are discarded show a small downward trend (not significant, with p-val=0.31 in a two-sided Kolmogorov–Smirnov test), but not as pronounced as in \textit{random document discard} method. Unlike the categorization test, the performance measure in the TOEFL Test present a high variability. This observation is consistent with the large fluctuations shown in \cite{Landauer1997}. Despite this, we consider it relevant to use this test to be able to compare with the results obtained by \cite{Landauer1997}.

\section*{Conclusion and Discussion}

Despite the popularity of word-embeddings in several semantic representation task, the way in which they acquire new semantic relations between words is unclear.
In particular, for the case of LSA there are two opposite visions about the effect of incorporating out-of-domain documents.
From one point of view, training LSA with a specific subcorpus, {\it cleaned} of documents unrelated to the specific task increases the performance \citep{Stone2006}.
From the other point of view, the presence of unrelated documents improves the representations. 
The second view point is supported by the conception that the SVD in LSA can capture high-order co-occurrence words relations \citep{Landauer1997,Lemaire2006,Turney2010}. 
Based on this, LSA is used as a plausible model of human semantic memory given that it can capture indirect relations (high-order word co-occurrences). 

In the present article we studied the effect of out-of-domain documents in LSA and Word2vec semantic representations construction. 
We compared two ways of progressive elimination of documents: the elimination of random documents vs the elimination of out-of-domain documents. 
The semantic representations quality was measured in two different tasks: a semantic categorization test and  a TOEFL exam. 
Additionally, we have varied a large range of word-embedding dimensions (\textit{k}). 

We have shown that Word2vec can take advantage of all the documents, obtaining its best performance when it is trained with the whole corpus. On the contrary, LSA's word-representation quality increases with a specialization of the training corpus (removal of out-of-domain document) accompanied by a decrease of \textit{k}. 
Furthermore, we have shown that the specialization without the decrease of \textit{k} can produce a strong performance reduction. 
Thus, we point out the need to vary \textit{k} when the corpus size dependency is studied.
From a cognitive modeling point of view, we point out that LSA's word-knowledge acquisitions does not take advantage of indirect learning (high-order word co-occurrences), while word2vec does. 
This throws light upon word-embeddings capabilities and limitations in modeling human cognitive tasks, such as: human word-learning \citep{Landauer1997,Lemaire2006,landauer2007}, semantic memory \citep{Denhiere2004,Kintsch2011,landauer2007} and  words classification \citep{Laham1997}.

\section*{Acknowledgments}
This research was supported by Consejo Nacional de Investigaciones Científicas y Técni-
cas (CONICET), Universidad de Buenos Aires, and Agencia Nacional de Promoción Científica y Tecnológica. We also want to thank LSA and NLP Research Labs, University of Colorado at Boulder for shearing access to the TOEFL word set. 

\bibliography{size_analysis}{} 
\appendix
\newpage
\section{Supplementary Materials: Latent Semantic Domain}
\begin{figure}[h!]
\begin{center}
\includegraphics[width=1\textwidth]{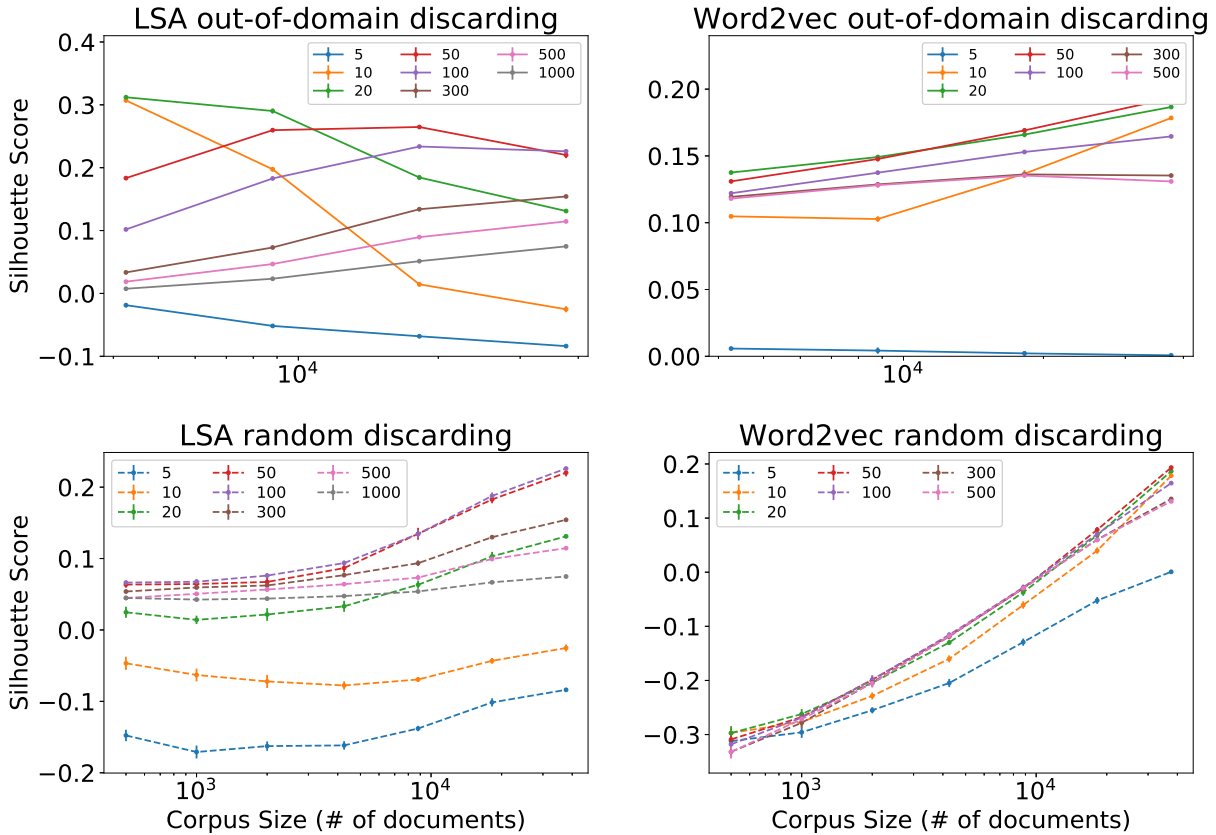}
\caption{ Semantic categorization test analysis. Categorization performance (Silhouette Score) vs corpus size, by number of  dimensions. Both document variation methods are shown: \textit{out-of-domain documents discarding} (top panels) and \textit{random document discarding} (bottom panels). The shown scores values and their error bars are, respectively, the mean values and the standard error of the mean of 10 samples.}
\label{S1_Fig}
\end{center}
\end{figure}

In figure \ref{S1_Fig} can be seen that performance decreases when documents are randomly discarded (bottom panels). However the dependency with out-of-domain documents (top panels) varied with the number of dimensions.
In the cases of 300, 500 and 100 dimensions, the performance decreases when out-of-domain documents are eliminated. 
In contrast, we obtain the opposite behavior in the cases of 5, 10, 20, 50, 100 dimensions, in which the elimination of out-of-domain documents increases LSA's categorization performance. 

Consider the case when \textit{k} is fixed in the value that maximizes the performance with the entire corpus (around $k=100$). 
When the corpus is ``cleaned'' of out-of-domain documents, the remaining corpus will have not only fewer documents, but also less topic diversity between texts. 
Thus, the number of dimensions (\textit{k}) needed to generate a proper semantic representation should be reduced. 
As $k$ is fixed in high values, LSA may not eliminate enough noisy dimensions, leading to a decrease in the performance. 
This effect becomes larger when the selected $k$ is higher, as it can be seen for $k=300$. 
On the other hand, consider the case when \textit{k} is fixed in the value that maximizes the performance with the ``cleaned'' corpus  (around $k=20$). 
The presence of out-of-domain documents in the complete corpus increase the topic diversity. 
As $k$ is fixed in low values, the LSA will not have enough dimensions to represent all the intrinsic complexity of the whole corpus. 
So, when the corpus is ``cleaned'' of out-of-domain documents, the performance should increase.

On the other hand, Word2vec present a performance decrease, with almost all dimension values, when out-of-domain documents are eliminated. Moreover, the discarding of out-of-domain documents do not require a considerable decrease of the number of LSA dimensions. These finding suggest that Word2vec do not encode \textit{latent semantic domains}, however more analysis must be done in these direction.

Unlike the categorization test, the performance measure in the TOEFL Test present a high variability. This observation is consistent with the large fluctuations shown in \cite{Landauer1997}. Despite this, we consider it relevant to use this test to be able to compare with the results obtained by \cite{Landauer1997}.

\begin{figure}[h!]
\begin{center}
\includegraphics[width=1\textwidth]{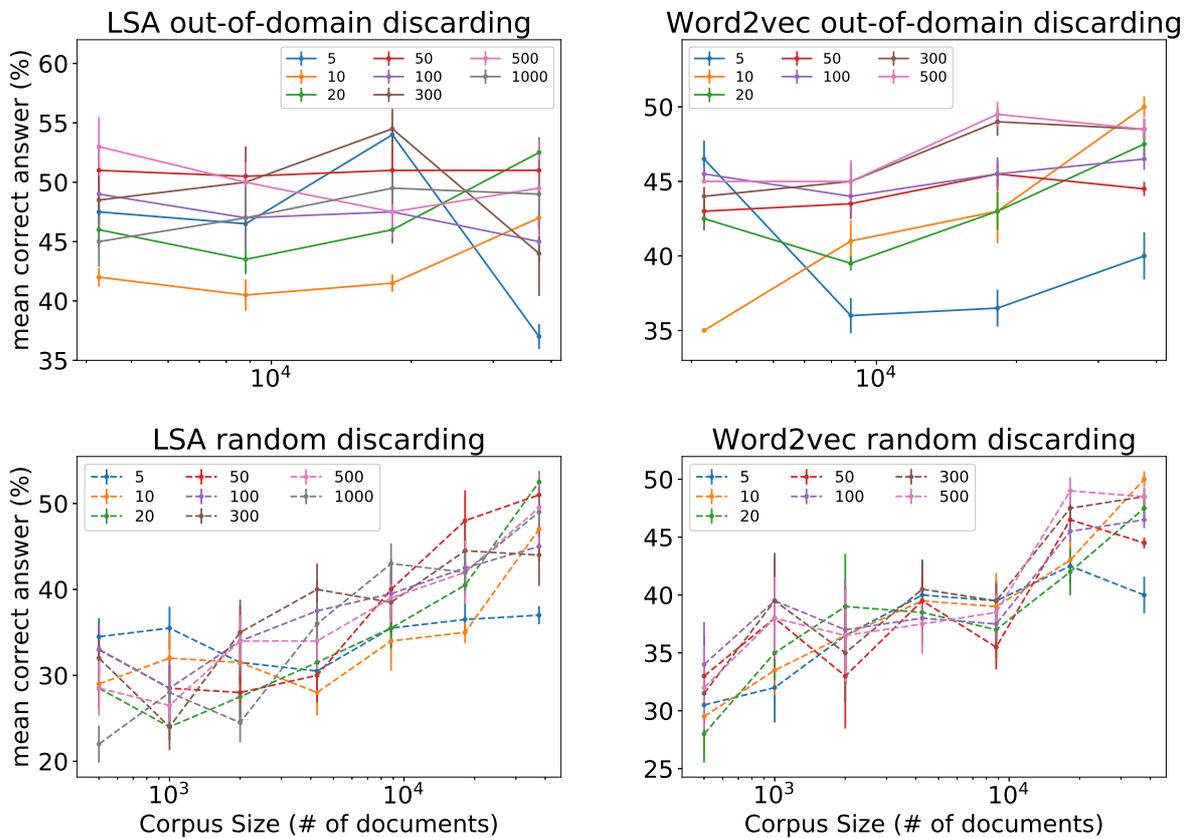}
\caption{TOEFL test analysis. Correct answer percentage vs corpus size, by number of  dimensions. Both document variation methods are shown: \textit{out-of-domain documents discarding} (top panels) and \textit{random document discarding} (bottom panels). The shown scores values and their error bars are, respectively, the mean values and the standard error of the mean of 10 samples.}
\label{S2_Fig}
\end{center}
\end{figure}
\end{document}